\documentclass[conference]{IEEEtran}
\IEEEoverridecommandlockouts
\usepackage[compatibility=false]{caption}

\usepackage{times}
\usepackage{epsfig}
\usepackage{graphicx}
\usepackage{amsmath}
\usepackage{amssymb}

\usepackage{xcolor}

\usepackage{booktabs} % Francisco
\usepackage{multirow} % Francisco%
\usepackage{hyperref} % Francisco
\usepackage{caption}
\usepackage{subcaption}

\def\BibTeX{{\rm B\kern-.05em{\sc i\kern-.025em b}\kern-.08em
    T\kern-.1667em\lower.7ex\hbox{E}\kern-.125emX}}

\begin{document}

%%%%%%%%% TITLE
\title{Evaluation of Few-Shot Learning Methods for Kidney Stone Type Recognition in Ureteroscopy}
%%%%%%%%% Authors------------------------------------------------------------------------------------
%%%%%%%%% ------------------------------------------------------------------------------------

\author{Carlos Salazar-Ruiz$^{\dagger,1}$,
Francisco Lopez-Tiro$^{\dagger,1,2}$, 
Ivan Reyes-Amezcua$^{1}$, \\
Clement Larose$^{2,3}$, 
Gilberto Ochoa-Ruiz$^{*,1}$,  
Christian Daul$^{*,2}$ \\
$^{1}$Tecnologico de Monterrey, School of Engineering and Sciences, Mexico \\
$^{2}$Université de Lorraine, CNRS, CRAN (UMR 7039), Vandœuvre-les-Nancy, France \\
$^{3}$CHRU de Nancy-Brabois, service d'urologie, Vandœuvre-les-Nancy, France 
\thanks{$^{\dagger}$Equal contribution. *Corresponding authors:}
\thanks{gilberto.ochoa@tec.mx, christian.daul@univ-lorraine.fr }
}

%%%%%%%%% ------------------------------------------------------------------------------------<

\maketitle

%%%%%%%%% ABSTRACT
\begin{abstract}

Determining the type of kidney stones is crucial for prescribing appropriate treatments to prevent recurrence. Currently, various approaches exist to identify the type of kidney stones. However, obtaining results through the reference ex vivo identification procedure can take several weeks, while in vivo visual recognition requires highly trained specialists. For this reason, deep learning models have been developed to provide urologists with an automated classification of kidney stones during ureteroscopies. Nevertheless, a common issue with these models is the lack of training data.
This contribution presents a deep learning method based on few-shot learning, aimed at producing sufficiently discriminative features for identifying kidney stone types in endoscopic images, even with a very limited number of samples. This approach was specifically designed for scenarios where endoscopic images are scarce or where uncommon classes are present, enabling classification even with a limited training dataset.
The results demonstrate that Prototypical Networks, using up to 25\% of the training data, can achieve performance equal to or better than traditional deep learning models trained with the complete dataset.

\end{abstract}

\section{Introduction}
\label{sec:intro}

\subsection{Medical context}

The formation of kidney stones in the urinary tract is a major public health issue \cite{lang2022global}. It has been reported that this condition affects between 10\% and 15\% of the world’s population. In the United States, 1 in 11 people has experienced an episode of kidney stones. Additionally, the risk of recurrence of the same type of stone has increased by up to 50\% \cite{daudon2018recurrence}.
The formation of kidney stones can be caused by a wide variety of factors, such as diet, a sedentary lifestyle, metabolic disorders, and low fluid intake \cite{daudon2004clinical}. Additionally, unavoidable factors like genetic predisposition, age, geographic region, climate, and chronic diseases increase the risk of kidney stone formation \cite{lang2022global}.
Therefore, methods for identifying different types of kidney stones are crucial for prescribing appropriate treatments and reducing the risk of recurrence \cite{daudon2018recurrence}.

Various procedures have been developed to identify kidney stones in clinical practice, such as the Morpho-constitutional Analysis (MCA) \cite{daudon2004clinical}, and more recently, Endoscopic Stone Recognition (ESR) \cite{estrade2022towards}.
MCA is the standard procedure for determining the different types of kidney stones (21 different types and subtypes, including both pure and mixed compositions) \cite{corrales2021classification}. MCA consists of a two-complementary analysis that combines the observation of morphology and a study of the biochemical composition of stones extracted from the urinary tract during ureteroscopy \cite{daudon2018recurrence}.
First, a morphological analysis is performed, where a biologist visually inspects the kidney stone using a magnifying glass. The goal of this inspection is to describe the stone in terms of color, texture, and morphology \cite{corrales2021classification}. This analysis is conducted for both the surface view (the external part of the kidney stone) and a cross-sectional view of the stone fragment (the internal part, which may consist of several layers surrounding the core). Subsequently, small fragments of the kidney stone are pulverized, and the resulting powder is used to analyze their biochemical composition through Fourier-transform infrared spectrophotometry (FTIR). The FTIR analysis provides a detailed description of the kidney stone’s biochemical composition \cite{daudon2004clinical}.
Finally, the MCA analysis provides the type of kidney stone through a detailed report on the biochemical and morphological characteristics of both views of the stone \cite{daudon2004clinical}. This technique is considered the current gold standard and has enabled the differentiation of up to 21 kidney stone subtypes.

However, MCA has some significant drawbacks: the results are usually available only after several weeks, and it is challenging to have the specialized team in every hospital needed to perform the MCA study.
For this reason, novel techniques such as Endoscopic Stone Recognition (ESR) have been developed to determine the most common types of kidney stones during the ureteroscopic intervention \cite{estrade2022towards}. The goal of such an approach is to identify the kidney stone type visually in real-time from the video feed on the screen. In this setting, the morphology of the surface and sections is analyzed and subsequently used by the urologist to confirm the lithogenesis. 
A recent study \cite{estrade2022towards} has shown that the visual recognition results performed by an expert on endoscopic images are highly correlated with the results obtained from morpho-constitutional analyses.
However, ESR requires a high degree of expertise due to the significant similarities between classes, and only a limited number of specialists possess this expertise. Additionally, this technique is highly operator-dependent and more subjective than MCA.

\subsection{Attempts at automating MCA and ESR}

Recent studies have demonstrated the ability of Deep Learning (DL) models to automatically classify kidney stones in both ex vivo and in vivo scenarios \cite{lopez2024vivo}.
These DL models have shown promising results and suggest great potential to assist urologists in making real-time decisions during ureteroscopy \cite{ali2022we}.
However, large amounts of data are required by these DL models to achieve accurate results. In the clinical context of kidney stones, collecting large datasets is a challenging task \cite{el2022evaluation}. The frequency with which the 21 subtypes of kidney stones appear can vary \cite{corrales2021classification}. For example, subtype “Ia” represents a higher frequency of occurrence (up to 18\%) compared to the less frequent subtype “VI” ($\le1\%$). For this reason, the number of samples per class in these datasets is highly imbalanced. In some cases, very few kidney stones of a specific subtype are represented in state-of-the-art datasets, which complicates the training of DL models, as they require a considerable amount of data for effective learning \cite{li2022trends}.

Although techniques for augmenting training data have been proposed \cite{lopez2021assessing} (which avoid training models from scratch), data scarcity and imbalance remain prevalent in the medical domain. In the state of the art, techniques such as Few-Shot Learning (FSL) have demonstrated remarkable results in image classification under scenarios with limited samples \cite{li2022trends}. FSL, in contrast to traditional methods, requires limited amounts of data to achieve high performance even with medical images.
The primary difference between FSL-based models and traditional DL models lies in how these models address the problem of data scarcity. FSL models are designed to be efficient in environments with limited data, while non-FSL models rely on large amounts of data to perform properly.

Given a scenario with limited samples, this work explores the use of Few-Shot Learning models to perform and improve kidney stone classification with a limited number of samples and compares their performance with traditional DL models. Additionally, ablation studies of FSL methods are presented to evaluate their efficiency under different configurations, such as the percentage of data used for training, backbone architecture, and the number of ways and shots.

This paper is organized as follows:
Section \ref{sec:material-methods} describes the construction of the dataset, reviews the key concepts of FSL methods, and outlines the FSL-based architectures evaluated in this work.
Section \ref{sec:result-discussion} compares the results obtained from the FSL architectures with their different configurations against traditional DL models.
Finally, Section \ref{sec:conclusion} analyzes future research directions.

\section{Materials and Methods}
\label{sec:material-methods}

\subsection{Dataset}
\label{sec:datasets}

For the experiments, an ex vivo endoscopic kidney stone dataset was used, as described in \cite{el2022evaluation}. The images were acquired using flexible, reusable digital ureteroscopes (i.e., endoscopes) from the Karl Storz brand. The dataset is described as follows:

% Jonathan El-Beze 
The ex vivo endoscopic dataset consisted of 409 images (see Table \ref{tab:dataset}). A total of 246 surface (SUR) images and 163 section (SEC) images were included. Six different kidney stone types were analyzed and categorized into subtypes denoted as WW (Whewellite, subtype Ia), WD (Weddellite, subtype IIa), UA (Uric Acid, subtype IIIa), STR (Struvite, subtype IVc), BRU (Brushite, subtype IVd), and CYS (Cystine, subtype Va). The images were captured using a phantom, in which kidney stone fragments were placed in an environment designed to realistically simulate the in vivo conditions of the urinary tract (for further details, refer to \cite{el2022evaluation}). All images had dimensions of 1920$\times$1080 pixels.

\begin{table}[t!]
\centering
\caption{Description of the endoscopic ex-vivo dataset \cite{el2022evaluation}. }

\vspace{-1mm}

\label{tab:dataset}
\resizebox{\columnwidth}{!}{%
\begin{tabular}{@{}ccccc@{}}
\cmidrule(r){1-5}
%\multicolumn{5}{c}{Dataset J. EL-BEZE et al. \cite{el2022evaluation}} \\ \cmidrule(r){1-5}
Subtype         & Main component (Key)         & Surface & Section & Total \\ \cmidrule(r){1-5}   \vspace{-0.5mm}
Ia              & Whewellite (WW)              & 62      & 25      & 87    \\  \vspace{-0.5mm}
IIa             & Weddelite (WD)               & 13      & 12      & 25    \\  \vspace{-0.5mm}
IIIa            & Uric Acid (UA)               & 58      & 50      & 108   \\  \vspace{-0.5mm}
IVc             & Struvite (STR)               & 43      & 24      & 67    \\  \vspace{-0.5mm}
IVd             & Brushite (BRU)               & 23      & 4       & 27    \\  \vspace{-0.5mm}
Va              & Cystine (CYS)                & 47      & 48      & 95    \\ \cmidrule(r){1-5}
\multicolumn{2}{c}{Number of images in dataset} & 246     & 163     & 409   \\ \cmidrule(r){1-5}  
\end{tabular}
}
\end{table}

\begin{figure}[t!]
  \centering
  \includegraphics[width=0.40\textwidth]{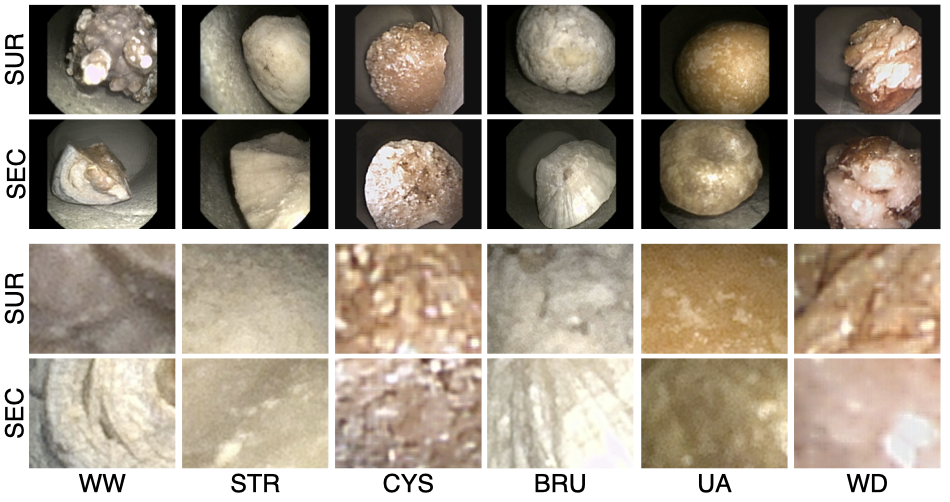}

  \caption{Examples of endoscopic kidney stone images (acquired ex-vivo). From top to bottom: Rows 1 and 2, surface and section images, respectively. Rows 3 and 4, 256$\times$256 patches from  rows 1 and 2, respectively.}
    \label{fig:dataset}
    \end{figure}

\begin{figure*}[h!]
  \centering
  \includegraphics[width=0.6\textwidth]{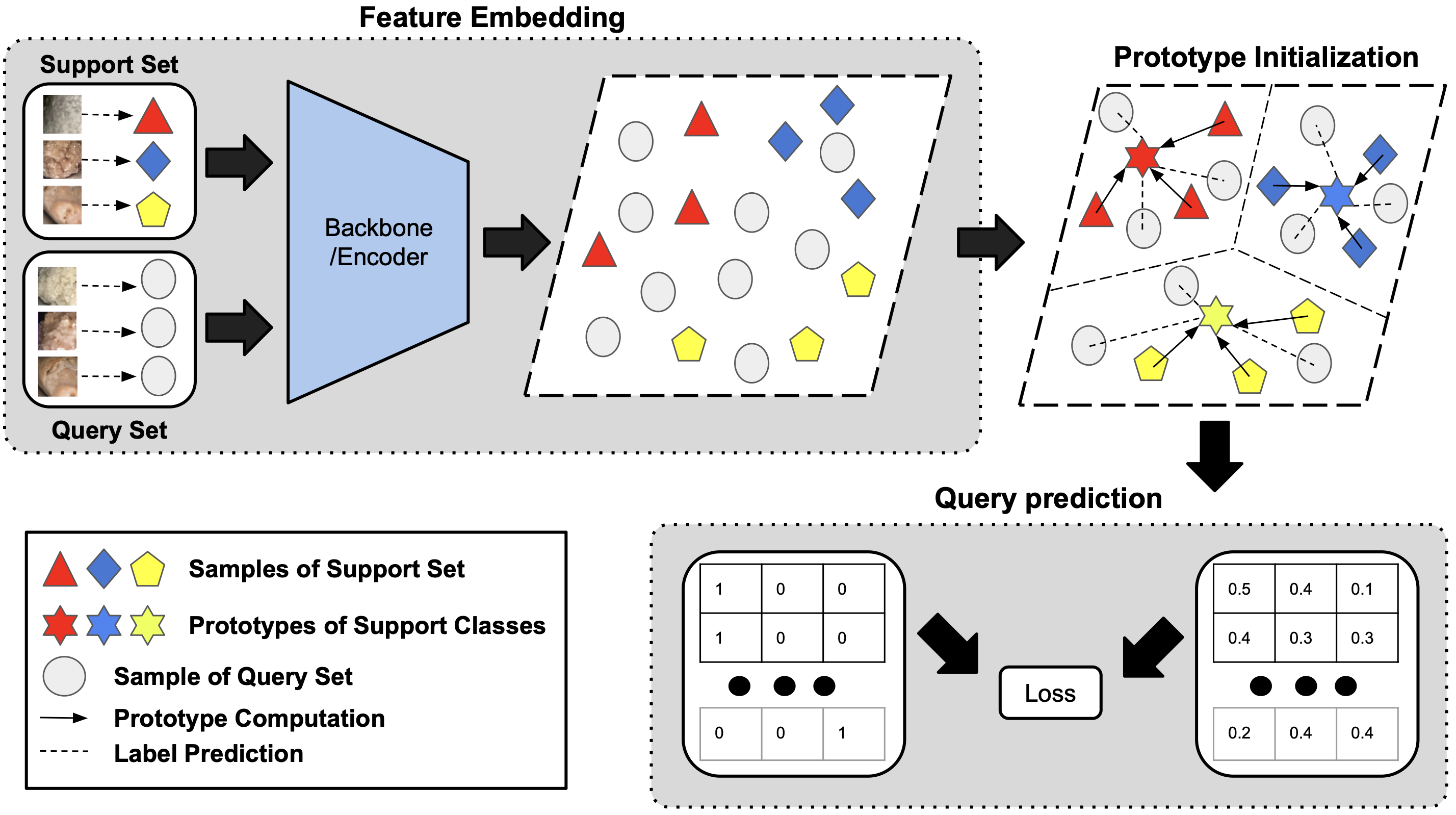}
 \vspace{-0.5mm}
  \caption{Representation of the Prototypical Networks method. Prototypical Networks is composed of three steps: feature embedding, prototype initialization, and query prediction. In the feature embedding stage, embeddings are extracted from the support set data using a backbone network employed as an encoder, such as ResNet. In the prototype initialization step, prototypes are generated from the labeled data in the support set using the extracted embeddings. Finally, during query prediction, the prototypes generated from the support set are compared to the features of the query set using a similarity metric.}
    \label{fig:ProtoNet}
    \end{figure*}

However, automatic kidney stone classification is generally not performed on full-resolution images due to the limited size of available datasets. Therefore, as in previous studies, 256$\times$256 pixel patches were extracted from the original images to increase and balance the dataset used for training (for further details, refer to \cite{lopez2024vivo}). The main advantage of using patches is that it allows deep learning models—which are particularly challenging to train with few samples—to be trained while simultaneously increasing the sample size and improving class balance.

In this study, three datasets for training and testing were created, corresponding to surface (SUR), cross-section (SEC), and mixed (MIX = SUR + SEC) views. A total of 6,000 patches were generated for both the SUR and SEC views. The MIX dataset was formed by combining the patches from SUR and SEC, resulting in a total of 12,000 patches. Each dataset was organized into six kidney stone subtypes, referred to as “classes” (see Figure \ref{fig:dataset}).
For experimentation purposes, each SUR, SEC, or MIX dataset was split such that 80\% of the patches were used for training and validation, while the remaining 20\% were used as test data. Patches extracted from the same image were included exclusively in either the training/validation set or the test set. Additionally, all patches were standardized using the mean $m_{i}$ and standard deviation $\sigma_{i}$ of the color values $I_{i}$ for each channel \cite{lopez2024vivo}.

\subsection{Few-Shot Learning}

Different deep learning (DL) approaches have demonstrated strong potential in recognizing various types of kidney stones, yielding promising results in both single views (SUR or SEC) and combined views (MIX) \cite{estrade2022towards, lopez2024vivo}. However, it remains difficult to obtain a large and balanced dataset in terms of class distribution. Consequently, different strategies have been implemented to increase the number of samples for underrepresented classes and to expand the dataset overall for DL training \cite{lopez2024vivo}. Traditional DL models trained on moderately large datasets have been shown to produce highly discriminative features across classes \cite{lopez2024vivo}. Nonetheless, the use of models capable of operating with a limited amount of data has not yet been thoroughly explored.

To address this limitation and extract meaningful information from a limited number of ex-vivo endoscopic images, this work proposes the use of Few-Shot Learning (FSL) models based on Prototypical Networks (ProtoNet) \cite{liu2022few} for kidney stone classification. Furthermore, rather than training the model from scratch, the ProtoNet backbone is initialized with weights pre-trained on ImageNet, which facilitates adaptation to the new data distribution.

\subsection{Prototypical Networks}
\label{sec:ProtoNet}

Image classification can be performed using various Deep Learning (DL) techniques. However, in scenarios with limited annotated data, Few-Shot Learning (FSL) models offer a promising alternative to address the challenges of data scarcity.
Prototypical Networks \cite{snell2017prototypical} are a DL-based FSL approach particularly suited for classification tasks where only a few labeled examples per class are available.

To illustrate the functioning of Prototypical Networks in the context of kidney stone classification, Figure~\ref{fig:ProtoNet} outlines the key stages of the pipeline. The model operates using two sets: a support set (used for learning) and a query set (used for evaluation). Both sets are passed through a feature extraction stage using a shared encoder or backbone network. For each class in the support set, a prototype is computed in the feature space—typically as the mean vector of its feature embeddings. Each query example is then classified based on its distance to these prototypes, assigning it to the closest one. The network is trained using a distance-based loss function (e.g., cross-entropy with Euclidean distances), which encourages separation between class prototypes in the embedding space.

In this work, we employ a one-step transfer learning strategy. Transfer Learning (TL) can be categorized into homogeneous and heterogeneous approaches. We adopt the heterogeneous TL (HeTL) paradigm, where the source and target domains differ. When only a small amount of training data is available, it is more effective to initialize the model with pretrained weights than with random values. Accordingly, we use ImageNet-pretrained ResNet models as backbones to extract feature embeddings for the Prototypical Networks. The overall workflow is divided into three main stages:

\subsubsection{Feature Embedding}
In this stage, both support and query samples are processed by the backbone network. The encoder—based on a ResNet architecture pretrained on ImageNet—projects the input images into a lower-dimensional embedding space. This transfer of knowledge helps improve generalization when training with limited data.

\subsubsection{Prototype Initialization}
After embedding extraction, the model computes one prototype per class by averaging the embeddings of the corresponding support samples. These class prototypes represent the centroids of each category in the feature space. This mechanism enables the model to generalize from just a few examples, as long as the embedding space captures discriminative information effectively.

\subsubsection{Query Prediction}
Each query sample is classified by computing its distance (typically Euclidean) to all class prototypes. The sample is assigned to the class with the closest prototype. During training, the model minimizes a cross-entropy loss based on these distances, encouraging embeddings of the same class to cluster around their prototype and different classes to be well-separated.

The goal of Prototypical Networks is to learn an embedding space where classes are distinct even with few examples. This capability is especially valuable in domains such as medical imaging, where acquiring large annotated datasets is costly or impractical. Additionally, the episodic training scheme—where the model is exposed to simulated few-shot classification tasks—helps prepare the model for real-world data scarcity scenarios.

\begin{table*}[!ht]
\centering
\caption{A performance comparison was conducted using Prototypical Networks with three different ResNet-based backbones (ResNet-18, ResNet-34, and ResNet-50) across the three kidney stone views: SUR, SEC, and MIX. Each performance metric (reported as mean $\pm$ standard deviation) corresponds to the average results obtained from experiments using a 6-way classification setting with varying numbers of shots (5, 10, 15, and 20) and training data proportions (100\%, 75\%, 50\%, and 25\%).  The best-performing result in each configuration is highlighted in bold.}
 \vspace{-1mm}
\label{tab:backbone}
\begin{tabular}{@{}ccccccc@{}}
\toprule
Method            & View         & Model             & Accuracy            & Precision           & Recall              & F1-Score            \\ \midrule \vspace{-0.0mm}
Prototypical Networks          & SUR          & ResNet18          & 84.68$\pm$2.39          & 85.62$\pm$2.06          & 84.68$\pm$2.39          & 84.60$\pm$2.32          \\ \vspace{-0.0mm}
\textbf{Prototypical Networks} & \textbf{SUR} & \textbf{ResNet34} & \textbf{86.65$\pm$2.22} & \textbf{87.56$\pm$2.04} & \textbf{86.65$\pm$2.22} & \textbf{86.51$\pm$2.17} \\ \vspace{-0.0mm}
Prototypical Networks          & SUR          & ResNet50          & 85.40$\pm$2.64          & 86.47$\pm$2.20          & 85.40$\pm$2.64          & 85.22$\pm$2.53          \\ \midrule \vspace{-0.0mm}
Prototypical Networks          & SEC          & ResNet18          & 90.96$\pm$3.10          & 91.48$\pm$2.85          & 90.96$\pm$3.10          & 90.88$\pm$3.14          \\ \vspace{-0.0mm}
\textbf{Prototypical Networks} & \textbf{SEC} & \textbf{ResNet34} & \textbf{92.86$\pm$1.93} & \textbf{93.37$\pm$1.74} & \textbf{92.98$\pm$1.87} & \textbf{92.93$\pm$1.90} \\ \vspace{-0.0mm}
Prototypical Networks          & SEC          & ResNet50          & 92.36$\pm$2.76          & 92.79$\pm$2.76          & 92.36$\pm$2.76          & 92.31$\pm$2.80          \\ \midrule \vspace{-0.0mm}
Prototypical Networks          & MIX          & ResNet18          & 87.20$\pm$1.72          & 87.89$\pm$1.61          & 87.24$\pm$1.63          & 87.14$\pm$1.68          \\ \vspace{-0.0mm}
\textbf{Prototypical Networks} & \textbf{MIX} & \textbf{ResNet34} & \textbf{87.98$\pm$1.76} & \textbf{88.52$\pm$1.55} & \textbf{87.98$\pm$1.76} & \textbf{87.97$\pm$1.78} \\ \vspace{-0.0mm}
Prototypical Networks          & MIX          & ResNet50          & 87.69$\pm$2.46          & 88.22$\pm$2.55          & 87.69$\pm$2.46          & 87.62$\pm$2.52          \\ \bottomrule \vspace{-0.0mm}
\end{tabular}
\end{table*}

\begin{table*}[ht]
\centering
\caption{Detailed performance comparison of Prototypical Networks based on the ResNet34 architecture (measured with accuracy). The best results for each configuration are denoted in bold.}
\label{tab:resnet34}

\begin{tabular}{@{}cccccccccr@{}}
\toprule
Method      & View & Backbone & Ways-Shots & 100\%          & 75\%           & 50\%           & 25\%        \\ \midrule \vspace{-0.50mm}
Prototypical Networks    & SUR  & ResNet34 & 6-5        & 86.70          & 86.77          & 85.62          & 84.85         \\ \vspace{-0.50mm}
Prototypical Networks    & SUR  & ResNet34 & 6-10       & \textbf{89.92} & \textbf{87.98} & 83.77          & \textbf{88.77}   \\ \vspace{-0.50mm}
Prototypical Networks    & SUR  & ResNet34 & 6-15       & 88.33          & 82.75          & \textbf{88.37} & 88.08           \\ \vspace{-0.50mm}
Prototypical Networks    & SUR  & ResNet34 & 6-20       & 88.33          & 85.67          & 82.65          & 87.88           \\ \vspace{-0.50mm}
\textcolor{black}{Traditional DL model} & \textcolor{black}{SUR}  & \textcolor{black}{ResNet34} & \textcolor{black}{--}         & \textcolor{black}{85.17}           & \textcolor{black}{86.00}           & \textcolor{black}{81.50}           & \textcolor{black}{77.00}           \\ \midrule \vspace{-0.5mm}
Prototypical Networks    & SEC  & ResNet34 & 6-5        & 91.13          & 94.02          & 90.78          & 91.42         \\ \vspace{-0.50mm}
Prototypical Networks    & SEC  & ResNet34 & 6-10       & \textbf{93.70} & \textbf{96.07} & \textbf{95.12} & 89.92          \\ \vspace{-0.50mm}
Prototypical Networks    & SEC  & ResNet34 & 6-15       & 92.87          & 94.62          & 90.93          & \textbf{95.22} \\ \vspace{-0.50mm}
Prototypical Networks    & SEC  & ResNet34 & 6-20       & 92.43          & 93.23          & 93.93          & 90.32          \\ \vspace{-0.50mm}
\textcolor{black}{Traditional DL model} & \textcolor{black}{SEC}  & \textcolor{black}{ResNet34} & \textcolor{black}{--}         & \textcolor{black}{91.75}           & \textcolor{black}{95.00}           & \textcolor{black}{90.50}           & \textcolor{black}{90.00}           \\ \midrule \vspace{-0.5mm}
Prototypical Networks    & MIX  & ResNet34 & 6-5        & 84.57          & 86.15          & 84.82          & 88.33          \\ \vspace{-0.50mm}
Prototypical Networks    & MIX  & ResNet34 & 6-10       & 87.17          & \textbf{89.63} & 87.42          & \textbf{90.17} \\ \vspace{-0.50mm}
Prototypical Networks    & MIX  & ResNet34 & 6-15       & \textbf{89.68} & 86.90          & \textbf{90.52} & 88.40          \\ \vspace{-0.50mm}
Prototypical Networks    & MIX  & ResNet34 & 6-20       & 88.90          & 87.47          & 88.75          & 88.77          \\ \vspace{-0.50mm}
\textcolor{black}{Traditional DL model} & \textcolor{black}{MIX}  & \textcolor{black}{ResNet34} & \textcolor{black}{--}         & \textcolor{black}{88.42}           & \textcolor{black}{87.06}           & \textcolor{black}{87.92}           & \textcolor{black}{86.17}           \\ 
\bottomrule \vspace{-0.50mm}
\end{tabular}
\end{table*}

\begin{figure*}[!ht] 
    \centering
    \subfloat[Prototypical Networks: Selected network ResNet34 (FSL) with 25\% of the dataset.]{\label{fig:prototypical}
\includegraphics[width=0.88\textwidth]{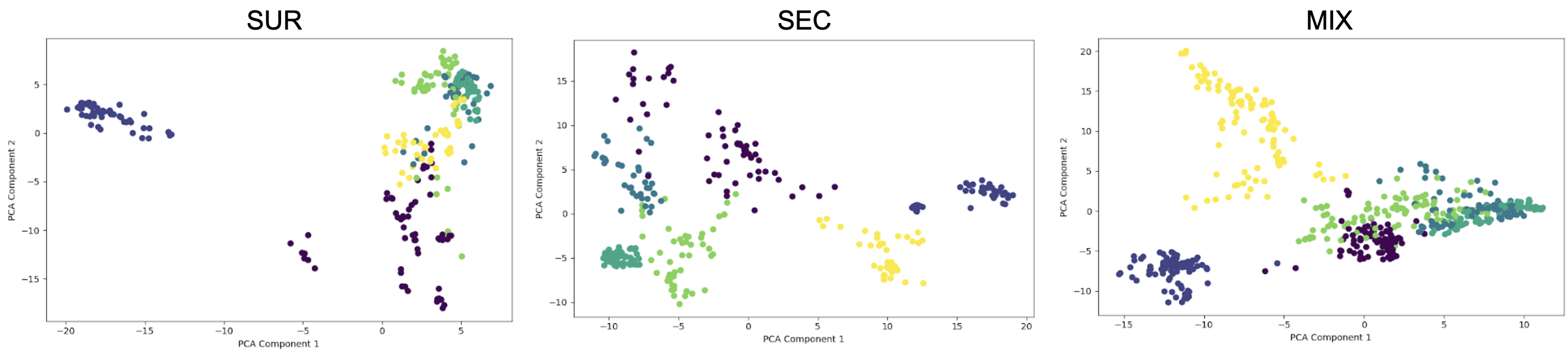}}
    \vspace{1mm}
    
    \subfloat[Traditional DL models: Selected network ResNet34 (non-FSL) with 100\% of the dataset.]{
\label{fig:traditional}\includegraphics[width=0.88\textwidth]{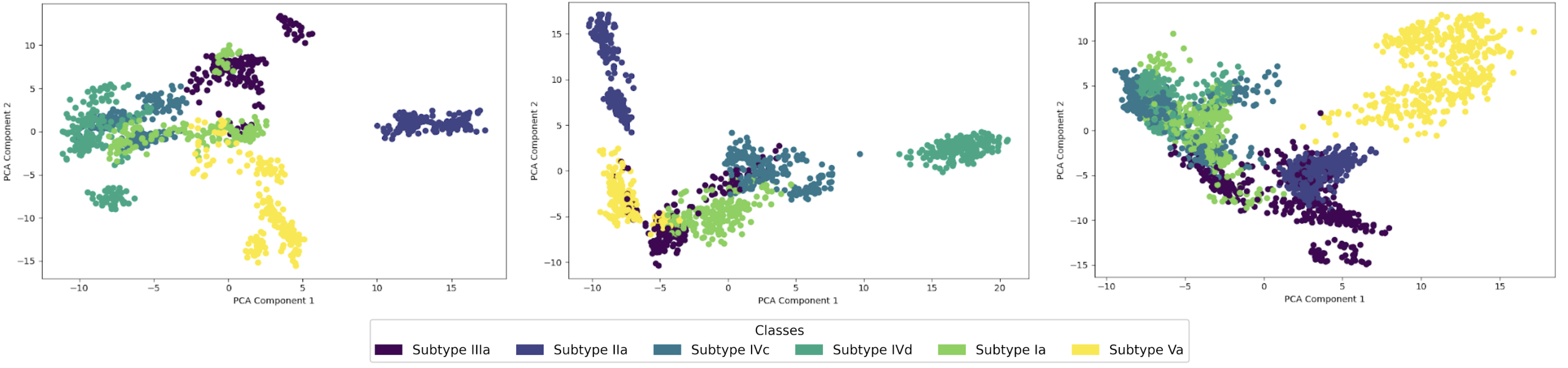}}

    \caption{Qualitative comparison between (a) Prototypical Networks using 25\% of the data and (b) traditional deep learning models (Traditional DL models) using 100\% of the data.} 
    \label{fig:tsne}
    \end{figure*}

\subsection{Experimental Configuration Setup}

To organize the experiments, the setup for training data, backbone configuration, and the Prototypical Networks “ways-shots” parameters are described. Additionally, implementation details are provided.

\subsubsection{Data Available for Training}
To train the model with a minimal amount of data, the dataset size was gradually reduced by randomly selecting subsets in 25% decrements. Model evaluation was performed using 100%, 75%, 50%, and 25% of the total dataset, ensuring that the experiments reflect the model’s performance across different amounts of training data.

\subsubsection{Prototypical Networks Configuration}
For training the ProtoNet-based model, the “ways” configuration was set to 6, corresponding to the number of classes in the dataset (see Table \ref{tab:dataset}). Additionally, the model’s performance was evaluated using 5, 10, 15, and 20 “shots” to test how the network handles varying numbers of labeled samples per class.

\subsubsection{Backbone Configuration}
The ResNet architecture was selected as the benchmark for this study, as it has been widely used in previous research to classify kidney stones into 4 and 6 categories with promising results \cite{estrade2022towards, lopez2024vivo}. However, determining the optimal network depth is not trivial, as the state of the art varies significantly in terms of the architecture used. In this study, three variants of ResNet—ResNet-18, ResNet-34, and ResNet-50—were tested to identify the most suitable configuration for the task at hand.

\subsubsection{Implementation Details}
For all experiments, we used PyTorch 2.5.1, torchvision 0.20.1, and the easyfsl 1.5.0 library for Few-Shot Learning experiments, with CUDA 12.4 for GPU acceleration. The backbones tested were ResNet-18, ResNet-34, and ResNet-50 from the torchvision model library. To obtain the feature vectors, the final fully connected layer of each ResNet model was replaced with a Flatten layer. The optimizer used was Adam with a learning rate of 0.0001. The model follows an episodic FSL approach, configured as “N Ways K Shots,” and was trained for 1000 iterations. The loss function employed was cross-entropy loss, evaluated over 100 iterations.

\section{Results and Discussion}
\label{sec:result-discussion}

Several experiments were conducted to evaluate Prototypical Networks, as described in Section \ref{sec:ProtoNet}, using the ex-vivo endoscopic dataset outlined in Section \ref{sec:datasets}. Specifically, the model’s ability to predict six different types of kidney stones across three views (SUR, SEC, and MIX) was assessed. Prototypical Networks was evaluated under four different “ways-shots” configurations (6-5, 6-10, 6-15, and 6-20), three backbone depth configurations (ResNet-18, ResNet-34, and ResNet-50), and four configurations of the percentage of data used for training (100\%, 75\%, 50\%, and 25\%).

\subsection{Prototypical Networks Results}
\label{sec:results-twostep}

The importance of analyzing models across different data scenarios lies in ensuring that Few-Shot Learning (FSL) models can correctly discriminate between classes regardless of the amount of data available (Fig. \ref{fig:prototypical}). These models should not only perform well with abundant data but also maintain robust performance when data is scarce.

To determine the best backbone for extracting features from each view, experiments were conducted across the four “ways-shots” configurations and the four arrangements of data available for training.
Table \ref{tab:backbone} presents a comparison of Prototypical Networks’ performance across these configurations. To calculate each performance (mean$\pm$std) shown in Table \ref{tab:backbone}, sixteen models were generated, based on different ways-shots configurations and the percentage of data used for training. For each performance measure (expressed as mean ± standard deviation), an overall average was computed across all ways-shots configurations, data percentages, and views (SUR, SEC, or MIX) as well as backbones (ResNet-18, 34, or 50).
As seen in Table \ref{tab:backbone}, the best overall performance for the SUR, SEC, and MIX views corresponds to ResNet-34. This architecture provides consistent performance across all metrics (accuracy, precision, recall, and F1-Score).
Although the MIX view yields the second-best performance with $87.98\pm1.76\%$ (accuracy), it is important to note that this performance was achieved using twice the data of the SUR or SEC views. In contrast, the performance of the SUR view ($86.65\pm2.22\%$ accuracy) is very similar to that of the MIX view.
Once ResNet-34 was selected, the goal was to determine the optimal “ways-shots” configuration and percentage of data. Table \ref{tab:resnet34} presents the results obtained with the ResNet-34 backbone.

The 6-ways-10-shots configuration consistently achieves the best performance across all views (SUR, SEC, or MIX) and various data percentages (100\%, 75\%, 50\%, and 25\%) using ResNet-34. However, when the 6-ways-10-shots configuration does not provide the best results, the 6-ways-15-shots configuration delivers the highest performance. While the 6-ways-20-shots and 6-ways-5-shots configurations do not achieve the best results, their performance is still very close to that of the 6-ways-10-shots and 6-ways-15-shots configurations.
Another interesting observation is that the performance (for any view: SUR, SEC, or MIX) achieved with 100\% of the available data for training is comparable to the performance achieved with 75\%, 50\%, or 25\% of the data. In other words, using less data still generates feature representations that are nearly as effective as those obtained with the full dataset. For instance, the performance of the 6-ways-10-shots configuration for the SUR view trained with 25\% of the data (88.77\% accuracy) is only slightly lower than that of the model trained with 100\% of the data (89.92\% accuracy). Similarly, for the 6-ways-10-shots configuration of the SEC view, the performance difference between using 100\% (93.70\%) and 25\% (89.92\%) of the data is just 3.78\% in terms of accuracy. Furthermore, there are configurations, such as the 6-ways-10-shots of the MIX view, where training with 25\% of the data (90.17\% accuracy) outperforms the full dataset (87.17\% accuracy).

\subsection{Comparison with Traditional DL Models}
\label{sec:comparison}

To evaluate the advantage of using Prototypical Networks compared to traditional deep learning models (Traditional DL models), we implemented a traditional ResNet-34 (without FSL) to classify the six classes for the SUR, SEC, and MIX views. The ResNet-34 models were trained using the same training data (100\%, 75\%, 50\%, and 25\%).

The results for the “Traditional DL models” are presented in Table \ref{tab:resnet34} and Fig. \ref{fig:traditional}. For the SUR view, training with 100\% of the data (2000 patches) results in a performance of 85.17\% using the traditional model (without FSL) and 89.92\% with ProtoNet. In a few-shot scenario, such as when using only 25\% of the data (500 patches), ProtoNet’s performance (88.77\%) remains superior to the traditional model’s performance (77.00\%).
For the SEC view, although the traditional DL model maintains similar performance (90.00\%) even when the data is reduced to 25\%, the ProtoNet model (95.22\%) continues to outperform it. Similarly, for the MIX view, the traditional model never surpasses the ProtoNet model in any scenario.

This demonstrates that using an FSL-based model, such as Prototypical Networks, is more efficient in the domain of kidney stone classification, especially when dealing with limited data.
The importance of analyzing models across different data scenarios lies in ensuring that FSL models can correctly discriminate between classes, regardless of the amount of data available. These models should perform well in situations with abundant data while maintaining good performance in cases with very limited data.

\section{Conclusion and Future Work}
\label{sec:conclusion}
\label{discussion_future_work}

This study demonstrates that Few-Shot Learning (FSL) methods, such as Prototypical Networks, enable the development of models that outperform traditional deep learning (DL) models. In particular, for kidney stone classification, FSL-based models maintain high performance even with a limited amount of data (only 25\% of the training set). However, further testing on other datasets is required to validate the effectiveness of these methods across different distributions, such as in-vivo endoscopic images and ex-vivo images captured with CCD cameras.

The results presented in this work suggest several directions for future research. One avenue is training models on full images rather than patches to explore the impact on performance. Additionally, expanding the comparison with other FSL models would be valuable to evaluate alternative approaches and identify the most effective methods for kidney stone classification.

\section*{Acknowledgements}
The authors acknowledge the support of the “Secretaría de Ciencia, Humanidades, Tecnología e Innovación” (SECIHTI), the French Embassy in Mexico, and Campus France through postgraduate scholarships, as well as the Data Science Hub at Tecnológico de Monterrey. This work was also funded by Azure Sponsorship credits from Microsoft’s AI for Good Research Lab under the AI for Health program and the French-Mexican Ecos Nord grant (MX 322537/FR M022M01).

{\small
\bibliographystyle{ieee_fullname}
\bibliography{references}

%\bibliography{references.bib}
%\bibliographystyle{ieee_full
}

\end{document}